\documentclass[10pt,twocolumn,letterpaper]{article}

\usepackage[pagenumbers]{cvpr} 

\usepackage[dvipsnames]{xcolor}

\definecolor{cvprblue}{rgb}{0.21,0.49,0.74}
\usepackage[pagebackref,breaklinks,colorlinks,citecolor=cvprblue,linkcolor=cvprblue,urlcolor=cvprblue]{hyperref}
\usepackage{multirow}
\usepackage{colortbl}
\usepackage{amssymb}
\usepackage{algorithm}
\usepackage{algpseudocode}

\newcommand{\mysize}{1.5in} 
 
\usepackage[labelformat=simple]{subcaption}

\title{Dynamic Erasing Network Based on Multi-Scale Temporal Features \\ for Weakly Supervised Video Anomaly Detection}

\author{Chen Zhang$^{1,2}$ \quad Guorong Li$^{3}$\thanks{Corresponding author.} \quad Yuankai Qi$^{4}$ \quad Hanhua Ye$^{3}$ \\ Laiyun Qing$^{3}$ \quad Ming-Hsuan Yang$^{5}$ \quad  Qingming Huang$^{1, 2, 3, 6}$\\
	$^{1}$State Key Laboratory of Information Security, Institute of Information Engineering, CAS \\
	$^{2}$School of Cyber Security, University of Chinese Academy of Sciences \\
	$^{3}$School of Computer Science and Technology, University of Chinese Academy of Sciences \\
	$^{4}$Australian Institute for Machine Learning, The University of Adelaide \\
    $^{5}$University of California, Merced \\
	$^{6}$Key Laboratory of Intelligent Information Processing, Institute of Computing Technology, CAS \\
	{\tt\small zhangchen@iie.ac.cn}, {\tt\small liguorong@ucas.ac.cn}, {\tt\small qykshr@gmail.com}, \\
	{\tt\small yehanhua20@mails.ucas.ac.cn},  {\tt\small lyqing@ucas.ac.cn}, {\tt\small mhyang@ucmerced.edu}, {\tt\small qmhuang@ucas.ac.cn}
}

\begin{document}
\maketitle
\begin{abstract}
The goal of weakly supervised video anomaly detection is to learn a detection model using only video-level labeled data. However, prior studies typically divide videos into fixed-length segments without considering the complexity or duration of anomalies. Moreover, these studies usually just detect the most abnormal segments, potentially overlooking the completeness of anomalies.  To address these limitations, we propose a Dynamic Erasing Network (DE-Net) for weakly supervised video anomaly detection, which learns multi-scale temporal features. Specifically, to handle duration variations of abnormal events, we first propose a multi-scale temporal modeling module, capable of extracting features from segments of varying lengths and capturing both local and global visual information across different temporal scales. Then, we design a dynamic erasing strategy, which dynamically assesses the completeness of the detected anomalies and erases prominent abnormal segments in order to encourage the model to discover gentle abnormal segments in a video. The proposed method obtains favorable performance compared to several state-of-the-art approaches on three datasets: XD-Violence, TAD, and UCF-Crime. Code will be made available at \href{https://github.com/ArielZc/DE-Net}{https://github.com/ArielZc/DE-Net}.

\end{abstract}    
\section{Introduction}
\label{sec:intro}

\begin{figure}[!t]
	\centering
	\includegraphics[width=\linewidth, trim=0 15 0 0,clip]{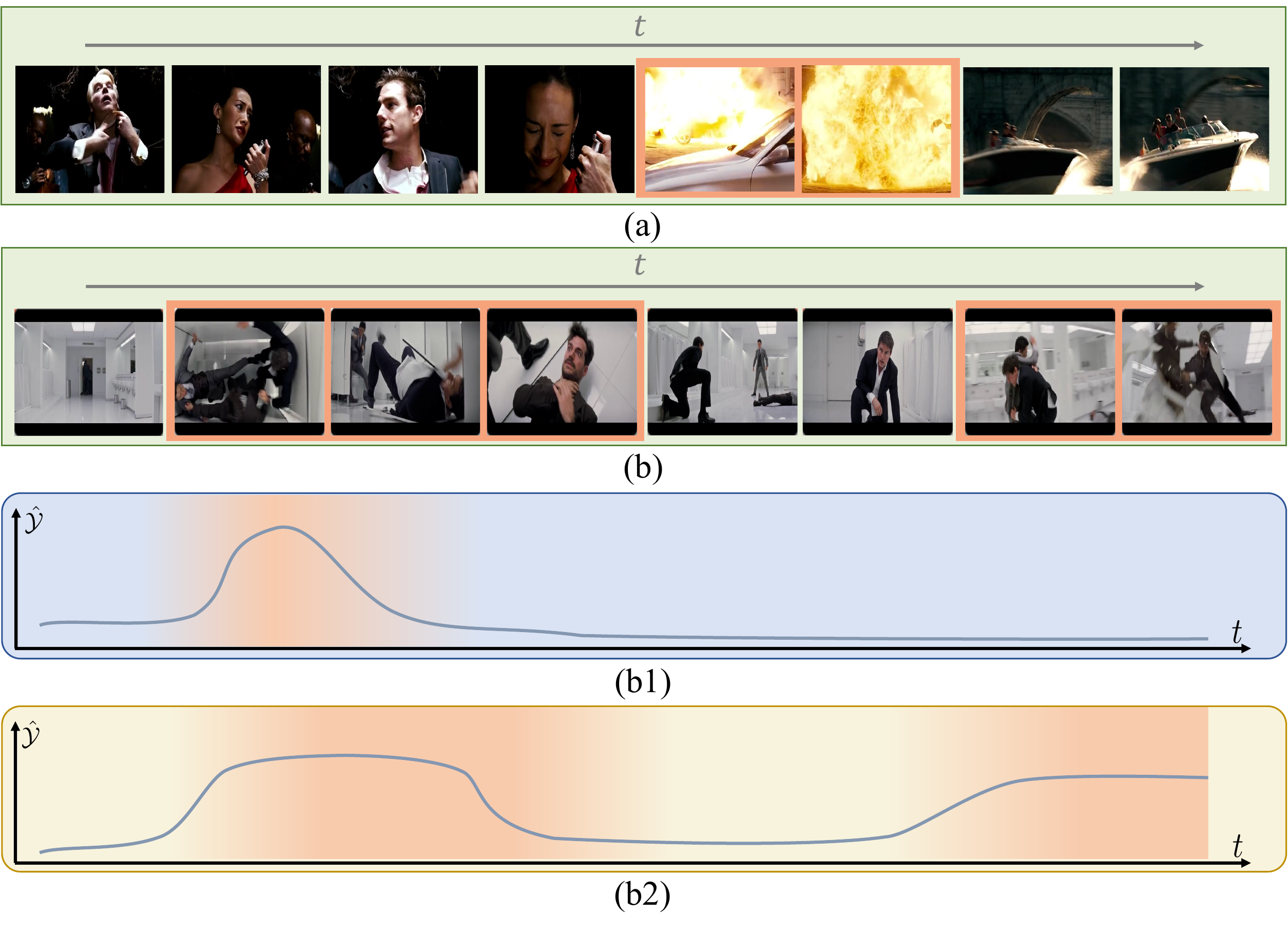}
	\caption{
		Illustration of our motivation. \textbf{Anomaly durations vary:} (a) A brief explosion anomaly, and (b) a fighting anomaly containing two non-contiguous fighting segments, one longer and the other shorter. \textbf{Completeness issue:} (b1) Conventional approaches detect only the most anomalous segments. (b2) Our method, using multi-scale temporal feature modeling and dynamic erasing, comprehensively detects long-term anomalies and non-contiguous abnormal segments.
	}
	\label{fig:motivation}
\end{figure}

Video anomaly detection aims to identify the abnormal events or actions in untrimmed videos.  To reduce the cost of collecting and labeling abnormal data, many semi-supervised learning methods ~\cite{xu2015learning, sabokrou2018adversarially, wu2019deep, park2020learning, cai2021appearance} and weakly supervised learning methods~\cite{zhang2019temporal, wan2020weakly, zaheer2020claws, pu2022locality, sapkota2022bayesian} have been proposed in recent years. 
Semi-supervised learning methods only require normal videos as training data and have witnessed significant achievements, but the performance of these methods is usually limited due to the lack of guidance from anomalous videos. Therefore,  weakly supervised learning methods that utilize abnormal videos with video-level annotations have gained more attention.
In this paper, we aim to further improve anomaly detection performance using weakly supervised learning.

In the weakly supervised setting, the training set is only given video-level annotations to indicate whether a video contains anomalies without detailed abnormal intervals. 
During the testing phase, the model needs to predict an anomaly score between 0 and 1 for each frame. Existing weakly supervised methods are mostly based on the Multiple Instance Learning (MIL) paradigm. Specifically, MIL-based methods~\cite{sultani2018real, zhang2019temporal, wan2020weakly, lv2021localizing} first divide each video into equal-length segments, where each video is viewed as a bag and each segment as an instance of the bag. 
Then, they utilize a ranking loss to maximize the margin between the anomaly scores of positive bags (abnormal videos) and negative bags (normal videos).
However, most MIL-based methods suffer from the following limitations: (I) Equal-length segments ignore the varying complexity or duration of anomalies. 
For instance, events like Explosion typically occur over a brief period, as shown in Fig.~\ref{fig:motivation}(a), whereas anomalies, such as Fighting, often last for a long duration.
Even for the same category of anomalies, as shown in Fig.~\ref{fig:motivation}(b), variations in duration are also present, with some being longer while others shorter. 
Dividing a video into a large number of segments (small temporal scale) provides fine-grained features but may lead to redundancy. 
Conversely, employing fewer segments (large temporal scale) yields coarser features but risks missing short-duration anomalies, which are likely to be overshadowed by numerous normal video frames. 
(II) Biased emphasis on the most abnormal segments. The ranking loss used in MIL-based methods always tends to detect the most unusual video segments, overlooking the remaining abnormal video segments. 
Certain anomalous events exhibit prolonged durations, and some videos encompass multiple discrete anomalies. 
Nevertheless, the models may excessively prioritize the most abnormal segments, thereby limiting their overall anomaly detection performance, as shown in Fig.~\ref{fig:motivation}(b1).

To address the above-mentioned issues, we propose a novel dynamic erasing network (DE-Net) based on multi-scale temporal features to mine complete abnormal events of various durations. 
Specifically, instead of dividing a video into a predefined number of segments,
we propose a Multi-Scale Temporal Modeling (MSTM) module to learn representations of segments of different lengths.
Through capturing and aggregating local and global visual information from video segments spanning different temporal scales,
our MSTM enables the acquisition of fine-grained as well as coarse-grained features, thereby enhancing the capacity to represent anomalies of diverse durations.
On the other hand, building upon the multi-scale temporal feature, we introduce a dynamic erasing strategy for the comprehensive detection of abnormal events.  
To adapt to the diversity of abnormal videos, this strategy dynamically evaluates the completeness of the detected anomalies, and discovers abnormal videos that need erasure operations. 
The dominant abnormal video segments in a video are erased to generate augmented abnormal videos, which are used to facilitate the model to discover mild anomalies.

The main contributions of this work are summarized as follows:
(1) We present a multi-scale temporal modeling module that diverges from the traditional single-scale approach, enabling dynamic adjustments in the number of video segments. 
(2) We propose a dynamic erasing strategy that mines complete anomalous segments. This strategy is characterized by its dynamic decision-making process, which determines when to apply erasure to the video and which segments should be erased, benefiting the uncovering of comprehensive anomalous segments. 
(3)  Experiments on three widely used datasets, XD-Violence, TAD, and UCF-Crime, demonstrate our method performs favorably compared to several state-of-the-art weakly supervised approaches.

\section{Related Work}
\label{sec:relatedwork}

\begin{figure*}[!t]
	\centering
	\includegraphics[width=0.95\linewidth]{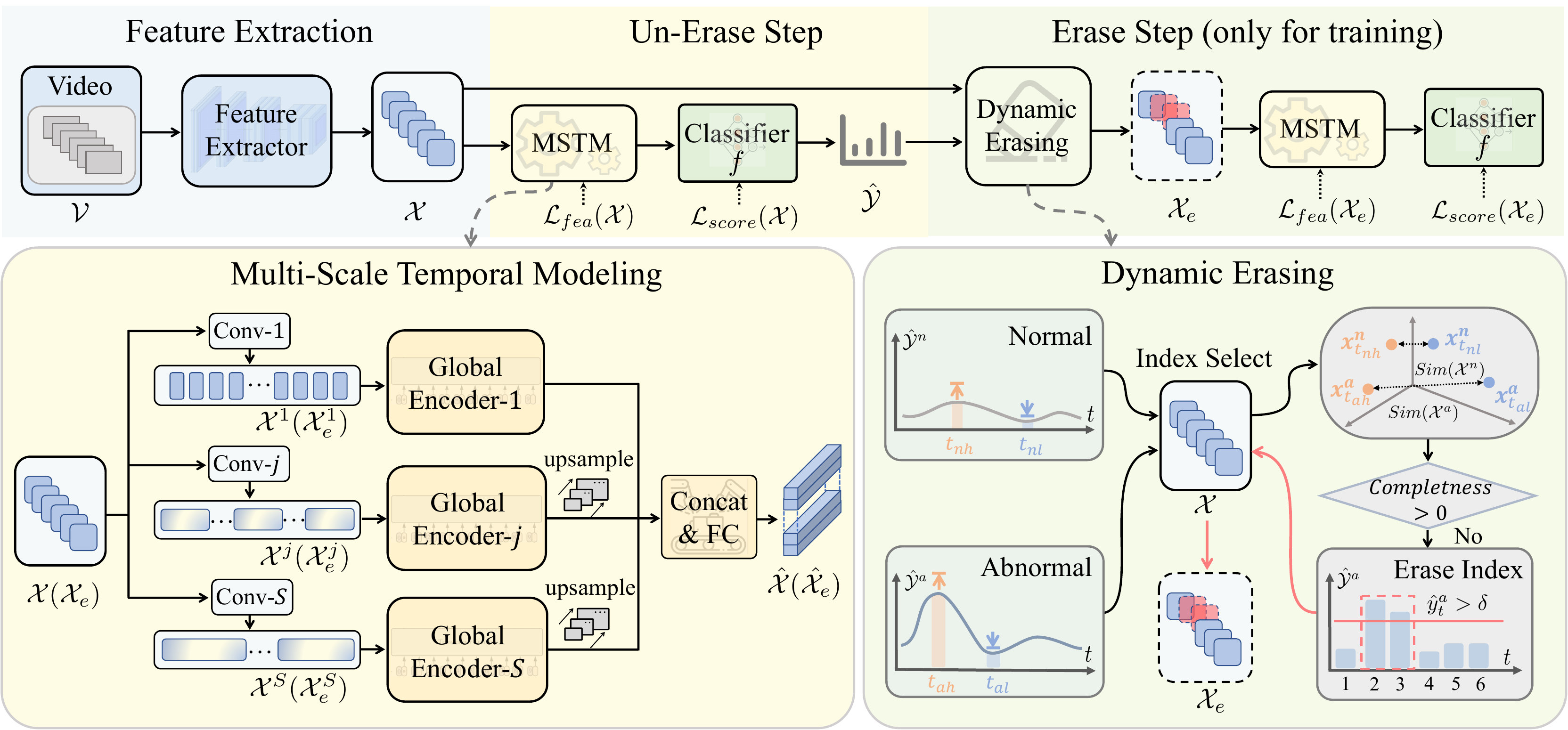}
	\caption{
		{Overview of our Dynamic Erasing Network (DE-Net) with multi-scale temporal features. The video $\mathcal{V}$ first goes through a pre-trained network to obtain original video features $\mathcal{X}$. 
			Then, in the Un-Erase Step, $\mathcal{X}$ is processed by the Multi-Scale Temporal Modeling module  (MSTM, Sec.~\ref{sec:MSTM}) to capture multi-scale local and global temporal features $\hat{\mathcal{X}}$. 
			Later, $\hat{\mathcal{X}}$ is fed into a classifier $f$ to predict anomaly scores $\hat{\mathcal{Y}}$.  
			Next, the Erase Step employs a Dynamic Erasing strategy (DE, Sec.~\ref{sec:de}), which utilizes $\hat{\mathcal{Y}}$ and segment feature similarities, to dynamically assess the completeness of detected anomalies in abnormal videos. 
			For videos whose detected anomalies are incomplete, segments with anomaly scores higher than $\delta$ are set to zeros, resulting in augmented video features ${\mathcal{X}_e}$.
			Then ${\mathcal{X}_e}$ is fed to the MSTM and Classifier again,
			to identify gentle abnormal segments. 
			In summary, we utilize original video features $\mathcal{X}$  and augmented video features ${\mathcal{X}_e}$ together for model optimization (Sec.~\ref{sec:model_train}).
		}
	}
	\label{fig:framework}
	\vspace{-1mm}
\end{figure*}

\noindent \textbf{Weakly Supervised Video Anomaly Detection. } 
Existing weakly supervised anomaly detection methods are mainly based on multiple instance learning framework~\cite{sultani2018real}, which employs ranking loss to enforce higher anomaly scores for segments within abnormal videos than those in normal videos.
These methods can be roughly divided into two types.
The first type attempts to improve the discriminative capacity between abnormal and normal features by designing more effective loss functions~\cite{zhang2019temporal, wan2020weakly, zaheer2020claws, pu2022locality, sapkota2022bayesian} or modeling temporal context relationships~\cite{zhu2019motion, wu2020not, wu2021learning, lv2021localizing, tian2021weakly, huang2022weakly, chen2023mgfn}. 
For example, in DDL~\cite{pu2022locality}, a dynamics ranking loss is introduced to amplify the magnitude of anomaly score changes, alongside a dynamics alignment loss that aligns the temporal feature dynamics and score dynamics within the video bag.  
In WSTD~\cite{huang2022weakly}, a transformer-styled temporal feature aggregator and a self-guided feature encoder are proposed to enhance the discriminative power of features. 
The second type of weakly supervised video anomaly detection methods~\cite{zhong2019graph, feng2021mist, li2022self, zhang2023exploiting} aims to convert the task into a supervised one, where clip-level pseudo labels are first generated and then fed into a classifier. 
For instance, Zhong \emph{et al.}~\cite{zhong2019graph} devise a graph convolutional network that corrects noisy labels by leveraging feature similarity and temporal consistency, thereby enhancing the quality of supervision for the classifier training. 
In~\cite{zhang2023exploiting}, a multi-head classifier is introduced to enhance the completeness of anomaly detection, but it maintains a fixed number of heads. In contrast, our approach employs a dynamic erasing strategy to selectively perform erasure operations for different videos, and thus becomes more flexible.

\vspace{1mm}
\noindent \textbf{Multi-Scale Temporal Feature Modeling.} Recent research in different fields, such as action detection~\cite{yang2021multi, dai2022ms} and person re-identification~\cite{li2020multi}, has shown the significance of multi-scale temporal feature modeling in enhancing video feature representation. For example, ConvTransformer is proposed in~\cite{dai2022ms} to model local and global temporal relations at multiple temporal scales. However, this network is designed for densely-labeled action detection; our work is rooted in weakly supervised video anomaly detection. 
In the field of anomaly detection, RTFM~\cite{tian2021weakly} employs multiple dilated convolutions to capture multi-scale local temporal dependencies. Differently, while modeling multi-scale local temporal features, our method also models global temporal relations at different scales.

\vspace{1mm}
\noindent \textbf{Erasing-based Methods.} The erasing strategy has been widely used in fields such as action detection~\cite{zhong2018step, zhang2019adversarial}, semantic segmentation~\cite{wei2017object, yoon2022adversarial}, and object detection~\cite{zhang2018adversarial, mai2020erasing} under weak supervision. 
The pipeline of such methods is to erase the prominent regions by setting a threshold and then mine less discriminative regions. However, these methods do not consider the differences in different videos or images. 
They erase the most discriminative areas for all samples, which is likely to cause over-erasure.
As far as the field of video anomaly detection is concerned, 
some videos contain only a few abnormal segments, while some videos contain much more abnormal segments, so different samples should be treated differently. 
Therefore, we design a dynamic erasing strategy based on segment feature similarity to dynamically and selectively perform erasure operations for each video.

\section{Method}
\label{sec:method}

In this section, we ﬁrst present the definition of the weakly-supervised video anomaly detection task; then, we briefly introduce the pipeline of the proposed Dynamic Erasing Network (DE-Net); at last, we provide details of our method.

\vspace{1mm}
\noindent\textbf{Task Definition.} Given a set of training videos $\left\{\mathcal{V}_i\right\}_{i=1}^N$ and its video-level label $\mathcal{Y}_i$ for each video $\mathcal{V}_i$, the goal of video anomaly detection is to build a model that can predict a set of frame-level anomaly scores for each test video, ranging from 0 to 1. 
Regarding the video label,  $\mathcal{Y}_i = 1$ if there is at least one abnormal clip in video $\mathcal{V}_i$; otherwise, $\mathcal{Y}_i = 0$.

\vspace{1mm}
\noindent\textbf{Framework Overview.} As shown in Fig.~\ref{fig:framework}, our DE-Net consists of two main steps: Un-Erase step and Erase step. Both steps exploit our Multi-Scale Temproal Modeling \textbf{MSTM} module (Sec.~\ref{sec:MSTM}), which aims to address the representation learning problem of abnormal events with different durations. In the Un-Erase step, the original video features are fed into MSTM and a classifier $f$ to predict their anomaly scores. 
In the Erase step, the Dynamic Erasing \textbf{DE} strategy (Sec.~\ref{sec:de}) leverages the predicted anomaly scores and feature similarities to predict which segments should be erased in the original features. After erasing,  the remaining features $\mathcal{X}_e$ 
are fed to MSTM and $f$ again for training, enabling the model to detect gentle abnormal segments.
In the inference stage, the features extracted from the test video are input into MSTM and $f$ to predict the anomaly score of each video frame.

\vspace{1mm}
\noindent\textbf{Initial Video Feature Extraction.} The initial video features are extracted 
as previous MIL-based methods ~\cite{tian2021weakly, wu2020not, wu2021learning}. Specifically, each video $\mathcal{V}_i$ is first divided into $T$ equal-length segments. Then, every 16 frames in each segment are viewed as a clip, and the pre-trained I3D network~\cite{carreira2017quo} is utilized to extract features from each clip. 
The averaged feature of clips is used to represent a segment. 
The segment features of video $\mathcal{V}_i$ are denoted as $\mathcal{X}_i=\left\{x_{i,1}, x_{i,2}, \ldots, x_{i,T}\right\}$, where $x_{i,t}$ represents the feature of the $t$-{th} segment. For simplicity, we will drop the video index $i$ and use $\mathcal{X}$ to represent the features of video $\mathcal{V}$ in the following. $\mathcal{X}^a$ and $\mathcal{X}^n$ are used to denote the features of abnormal video and normal video, respectively.

\subsection{Multi-Scale Temporal Modeling (MSTM)}
\label{sec:MSTM}

For each video, we first learn segment features of varying lengths at different temporal scales. 
These diverse scale features are then used for global temporal modeling. 
At last, we fuse these temporal features to comprehensively leverage both local and global contextual information across different scales.

\vspace{1mm}
\noindent \textbf{Multi-Scale Local Temporal Modeling.}   To obtain video segment features of different temporal lengths, we apply several 1D convolutional layers with different strides and kernel sizes (as shown in Fig.~\ref{fig:framework}) on the initial segment features $\mathcal{X}=\left\{x_{1}, x_{2}, \ldots, x_{T}\right\} \in \mathbb{R}^{T \times D}$ along the time dimension:
\begin{equation}
\bar{\mathcal{X}}^s = \operatorname{Conv1D}(\mathcal{X}, *^{(s)}), s \in [1, 2, ..., S],
\end{equation}
where $*^{(s)} = (stride=2^{s-1}, kernel\_size=2^{s-1})$  represents the convolution operator at scale $s$.
The notation $\bar{\mathcal{X}}^s = \{\bar{x}_{1}^s, \bar{x}_{2}^s, \ldots, \bar{x}_{\frac{T}{2^{s-1}}}^s\} \in \mathbb{R}^{\frac{T}{2^{s-1}} \times D} $ denotes the learned video feature sequence at scale $s$, and $\frac{T}{2^{s-1}}$ is the new temporal length of the convolution output.
In this way, we obtain multi-scale local temporal features and capture information between neighboring segments. 

\vspace{1mm}
\noindent \textbf{Impose Global Awareness.} For local temporal features $\bar{\mathcal{X}}^s $ at scale $s$, we propose to learn its global temporal representation via  transformer~\cite{vaswani2017attention}, where each segment feature $\bar{x}_{t}^s$ is viewed as a token.  
To indicate temporal order among tokens, we apply a learnable positional encoding (PE) to these tokens. Then, local temporal features and position embeddings are fed to the transformer backbone comprised of Multi-head Self Attention (MSA), Layer Norm(LN), and Multi-Layer perceptron (MLP):
\begin{equation}
\hat{\mathcal{X}}^s= \operatorname{TransEncoder^s}(\bar{\mathcal{X}}^s+\operatorname{PE}^s).
\end{equation}

\noindent \textbf{Feature Aggregation.} To better represent abnormal events of diverse durations, we further fuse the multi-scale global-aware local features ${\mathcal{X}^{MS} }= \{\hat{\mathcal{X}}^s\}_{s=1}^S$.
As these features are of different lengths, we first align them to the same length $T$ via nearest neighbor linear interpolation:
\begin{equation}
\hat{\mathcal{X}}^s= \begin{cases}\hat{\mathcal{X}}^{s}, &  \text { if } s=1 \\ 
\{\operatorname{Repeat}(\hat{x}_t^s, 2^{s-1})\}_{t=1}^{\frac{T}{2^{s-1}}} , & \text { otherwise }\end{cases}
\end{equation}
where $\operatorname{Repeat}(\hat{x}_t^s, 2^{s-1})$ indicates that we repeat $2^{s-1}$ times for each segment feature $\hat{x}_t^s$.
Then, we concatenate them and pass the concatenated features through an FC layer to learn the final video feature representation $\hat{\mathcal{X}}=\left\{\hat{x}_{1}, \hat{x}_{2}, \ldots, \hat{x}_{T}\right\} \in \mathbb{R}^{T \times D}$:
\begin{equation}
\hat{\mathcal{X}} =  \operatorname{FC} (\operatorname{Concat}(\hat{\mathcal{X}}^1, \hat{\mathcal{X}}^2 \ldots,\hat{\mathcal{X}}^S )),
\end{equation}
which is later fed into the classifier $f$ for anomaly classification. The classifier $f$ is composed of three fully connected layers, with 512, 32, and 1 units in each layer. Dropout regularization with a probability of 0.6 is applied between each layer to prevent overfitting. ReLU activation function is used after the first layer, while the final layer utilizes a Sigmoid activation function. 

\subsection{Dynamic Erasing (DE)}
\label{sec:de}

Anomaly detection models usually risk detecting only the prominent anomalous segments, rather than all anomalous segments.
To address this problem,
we design a dynamic erasing strategy that first dynamically assess the completeness of detected anomalies in each abnormal video, and then erases prominent abnormal segments from videos with incomplete anomaly detection, forcing the model to discover more anomalous segments.

\vspace{1mm}
\noindent \textbf{Dynamic Assessment.} Since the abnormal videos vary largely, the erasure operation for distinct abnormal videos should be video-specific. 
It is noteworthy that the similarity between abnormal segments and normal segments within an abnormal video tends to be small, while the similarity between segments within a normal video is generally high. Therefore, we propose to measure the completeness of the detected anomalies based on the difference between the segment similarity within the abnormal video and the segment similarity within the normal video.

Specifically, we first get the predicted anomaly scores $\hat{\mathcal{Y}} $ by feeding the multi-scale features $\hat{\mathcal{X}}$ into the classifier $f$:
\begin{equation}
\hat{\mathcal{Y}} =  f(\hat{\mathcal{X}}),
\end{equation}
where $\hat{\mathcal{Y}} = \{\hat{y}_{1}, \hat{y}_{2}, \ldots, \hat{y}_{T}\} \in \mathbb{R}^{T \times 1}$ and $ \hat{y}_{t}$ denotes the anomaly score of the $t$-{th} segment of the video.
Then, we assume that the segment with the highest anomaly score is the most anomalous segment and the segment with the lowest anomaly score is the most normal segment. We apply max and min selection on the temporal dimension of anomaly scores to find the most abnormal and normal segment index in the videos:
\begin{equation}
{t}_h =  \operatorname{argmax}(\hat{\mathcal{Y}}), \;
{t}_l =  \operatorname{argmin}(\hat{\mathcal{Y}}),
\end{equation}
where ${t}_h$ and ${t}_l$ represent the index of the most abnormal and normal segments, respectively. For the normal video $\mathcal{X}^n$, we use ${t}_{nh}$ and ${t}_{nl}$ to denote the corresponding indexes, and use ${t}_{ah}$ and ${t}_{al}$ for the abnormal video $\mathcal{X}^a$. 

Next, we calculate the similarity between the most abnormal segment and the most normal segment in a video via
\begin{equation}
Sim(\mathcal{X}) = \frac{x_{t_{h}}^{\top} x_{t_{l}} }{\left\|x_{t_{h}}\right\|\left\|x_{t_{l}}\right\|}.
\end{equation}
Generally, $Sim(\mathcal{X}^a)$  is smaller than $Sim(\mathcal{X}^n)$. 
For an abnormal video, a smaller $Sim(\mathcal{X}^a)$ means there is a larger difference between the detected abnormal  and normal segments, indicating the detected abnormal segments may be incomplete. 
Thus, we define the completeness of the detected anomalies in the abnormal video $\mathcal{X}^a$ as
{\small
    \begin{equation}
        Completeness(\mathcal{X}^a) \! = \! Sim(\mathcal{X}^a) \! - \!  \frac{1}{|B^n|} \!  \sum_{\mathcal{X}^n \in B^n} \! \! \!  Sim(\mathcal{X}^n),
    \end{equation}}
$\! \!$where $B^n$ represents the set of normal videos in the training batch. 
If $Completeness(\mathcal{X}^a)$ is larger than 0, we consider there are no further abnormal segments within the abnormal video, and thus, we do not perform erasure operation on the video. Otherwise, we erase the significant abnormal segments from the video.
Concretely, we employ an erase memory to record whether the abnormal video $\mathcal{X}^a$ needs to perform erasure operation as follows:
\begin{equation}
EraM(\mathcal{X}^a)= \begin{cases}0, & \text { if } Completeness (\mathcal{X}^a)> 0 \\ 
	1 , & \text { otherwise }\end{cases}
	\end{equation}
	where 0 means not to perform the erasure operation, and 1 means the opposite.

\vspace{1mm}
\noindent \textbf{Erasure Operation.} For the abnormal video that requires erasure operation, we erase the prominent abnormal segments in the temporal dimension.
Specifically, we set a score threshold to determine which segments in the video are significant anomalies and set these segment features as zeros. The erasure operation on the original abnormal video features $\mathcal{X}^a=\{x_t^a\}_{t=1}^T$ is formulated as follows:
	\begin{equation}
		\tilde{x}_t^a= \begin{cases}0, & \text { if } \hat{y}_t^a > \delta \  and \ EraM(\mathcal{X}^a) =1 \\ 
			x_t^a , & \text { otherwise }\end{cases}
	\end{equation}
	where $\hat{y}_t^a$ denotes the anomaly score of the segment $x_t^a$, and $\delta$ represents the erase threshold. 
	After performing the above erasure operation, we obtain the erased  features ${\mathcal{X}_e}=\{\tilde{x}_t^a\}_{t=1}^T$, 
	which are subsequently treated as augmented abnormal video features and fed to the network again. Thus the model can be forced to detect more abnormal segments.
	

\subsection{Model Optimization}
\label{sec:model_train}
   As in previous work~\cite{sultani2018real}, we model anomaly detection as a regression problem.  To encourage the anomalous video segments to have higher anomaly scores than the normal segments, we adopt the hinge-based MIL ranking loss to maximize the gap between the highest anomaly score in the abnormal video and the highest anomaly score in the normal video:
	\begin{equation}
\mathcal{L}_{score}=\max \left(0,1-\max _{\hat{y}_{t}^a \in \hat{\mathcal{Y}}^a} \hat{y}_{t}^a +\max _{\hat{y}_{t}^n \in \hat{\mathcal{Y}}^n} \hat{y}_{t}^n \right),
\end{equation}
where $\max$ is computed across all video segments within each video, $\hat{\mathcal{Y}}^a$ and $\hat{\mathcal{Y}}^n$ denote the anomaly scores of the abnormal video and normal video, respectively. 
In detail, $\hat{\mathcal{Y}} = \{\hat{y}_{1}, \hat{y}_{2}, \ldots, \hat{y}_{T}\} \in \mathbb{R}^{T \times 1}$  and $\hat{y}_{t}$ denotes the anomaly score of the $t$-{th} segment in the video. 

\begin{algorithm}[!t]
	\caption{Training Process of DE-Net}
	\label{alg:de-net}
	\begin{algorithmic}[1]
		\Require Video data $\mathcal{V}$
		\Ensure DE-Net
		\State Extract video features: $\mathcal{X} \gets \operatorname{FeatureExtractor}(\mathcal{V})$
		\While{ training is not converged }
		\State // Un-Erase Step
		\State Get multi-scale features $\hat{\mathcal{X}} \gets \operatorname{MSTM} (\mathcal{X})$
		\State Generate anomaly scores $\hat{\mathcal{Y}} =  f(\hat{\mathcal{X}})$
		\State // Erase Step
		\State Obtain erased video features ${\mathcal{X}_e} \gets \operatorname{DE}(\mathcal{X, \hat{\mathcal{Y}}})$
		\State Get multi-scale features $\hat{\mathcal{X}_e} \gets \operatorname{MSTM}(\mathcal{X}_e)$
		\State Generate anomaly scores $\hat{\mathcal{Y}_e} =  f({\mathcal{X}_e})$
		\State Calculate unerased loss 	$\mathcal{L}_{u}$ by Eq. (\ref{eq:loss_unera})
		\State Calculate erased loss $\mathcal{L}_{e}$ by Eq. (\ref{eq:loss_era})
		\State Calculate the total loss  $\mathcal{L}$ by Eq. (\ref{eq:loss_total})
		\State Update the parameters of $\operatorname{MSTM}$ and $f$
		\EndWhile
		\State \textbf{return} DE-Net
	\end{algorithmic}
\end{algorithm}

For the final multi-scale video temporal feature  $\hat{\mathcal{X}}=\left\{\hat{x}_{1}, \hat{x}_{2}, \ldots, \hat{x}_{T}\right\}$, we measure the variation in features between the current segment $\hat{x}_t$ and its $k$-{th} temporal neighbor $\hat{x}_{t-k}$ by first calculating the cosine similarity: $\operatorname{cos}(\hat{x}_{t-k}, \hat{x}_{t})  = \hat{x}_{t-k}^{\top} \hat{x}_t /(\left\|\hat{x}_{t-k}\right\|\left\|\hat{x}_{t}\right\|)$, where $k=T/2$.  Then,  the local variation of segment $\hat{x}_t$  is computed as:
	\begin{equation}
		var(\hat{x}_t)= 1 - \operatorname{cos}(\hat{x}_{t-k}, \hat{x}_{t}).
	\end{equation}
From the feature level, it is encouraged that the variation in abnormal video should also be larger than that in normal video:
\begin{equation}
    \mathcal{L}_{fea} \! = \! \max \! \left(\! 0,1 \! - \! \! \! \max_{\hat{x}_{t}^a \in \hat{\mathcal{X}}^a} \! var(\hat{x}_{t}^a) \! + \! \!  \max _{\hat{x}_{t}^n \in \hat{\mathcal{X}}^n} \! var(\hat{x}_{t}^n) \! \right) \! . \!
\end{equation}

When erasure is not performed, the loss function of the network is as follows:
\begin{equation}
\mathcal{L}_{u} = {\alpha}_1 \mathcal{L}_{score}({\mathcal{X}}) + {\alpha}_2 \mathcal{L}_{fea}({\mathcal{X}}).
\label{eq:loss_unera}
\end{equation}
After erasing the prominent abnormal segment features, the loss becomes to:
\begin{equation}
\mathcal{L}_{e} = {\alpha}_1 \mathcal{L}_{score}({\mathcal{X}_e}) + {\alpha}_2 \mathcal{L}_{fea}({\mathcal{X}_e}),
\label{eq:loss_era}
\end{equation}
where $\alpha_1$ and $\alpha_2$ are hyper-parameters to balance the contribution of each loss function. 
Finally, the overall loss function of the entire dynamic erasing network is:
\begin{equation}
\mathcal{L} = 	\lambda_1 \mathcal{L}_{u} +  \lambda_2 \mathcal{L}_{e},
\label{eq:loss_total}
\end{equation}
where $\lambda_1$ and $\lambda_2$ are hyper-parameters to balance the unerased loss and erased loss.  
In summary, Algorithm~\ref{alg:de-net} depicts the complete training procedure of our Dynamic Erasing Network (DE-Net).

\section{Experiments}

In this section, we first introduce the experimental settings. Then, we present quantitative comparison results with the state-of-the-art methods on three datasets: XD-Violence~\cite{wu2020not}, TAD~\cite{lv2021localizing}, and UCF-Crime~\cite{sultani2018real}.  Last, we provide extensive ablation studies to analyze the effect of each design choice in our model.

\subsection{Experimental Settings}

\noindent \textbf{Datasets.} (I) XD-Violence consists of 3954 training videos annotated at the video level and 800 test videos featuring frame-level annotations, which is the largest dataset for anomaly detection. 
Moreover, it is the most challenging dataset due to that the videos often contain camera movements and scene switching. 
(II) TAD is a dataset consisting of traffic videos that are mostly recorded by CCTV cameras mounted on the roads. 
It encompasses seven type of real-world anomalies, distributed across 400 training videos and 100 test videos. Following the weak supervision paradigm, the training set is annotated with video-level labels, while the test set offers annotations at the frame level. 
(III) UCF-Crime is a large-scale dataset comprised of real-world videos captured by surveillance cameras. It consists of 1,610 training videos annotated with video-level labels and 290 test videos annotated at the frame level to facilitate performance evaluation. The videos are collected from different scenes and encompass 13 distinct categories of anomalies.

\vspace{1mm}
\noindent \textbf{Evaluation Metrics.} For the XD-Violence dataset, we adopt Average Precision (AP) to evaluate the model as in~\cite{tian2021weakly, wu2020not, wu2021learning, li2022self}. Following previous models~\cite{sultani2018real, zhong2019graph, feng2021mist}, we use the frame-level area under the curve (AUC) of the ROC curve as the evaluation metric for the UCF-Crime and TAD datasets. Higher AUC and AP values denote better performance.

\vspace{1mm}
\noindent \textbf{Implementation Details.} For clip-level features, we follow~\cite{tian2021weakly} to use the Kinetics pre-trained I3D network~\cite{carreira2017quo} to extract features, where the input is a video clip consisting of 16 consecutive frames. 
We also use the VGGish network~\cite{gemmeke2017audio} to extract audio features for the XD-Violence dataset as in~\cite{wu2020not}. The segment number $T$ is set as $64$. For the multi-scale temporal features modeling, the scale number $S$ is set as $3$.  We implement a one-layer transformer encoder for global temporal modeling, and the head number is set to $8$. 
For the dynamic erasing strategy, we set the erase threshold $\delta$ as $0.8$. For the loss function, $\alpha_1=1.0$, $\alpha_2=1e-4$, and $\lambda_1=\lambda_2=1.0$. To train the whole network, we adopt the Adam optimizer with a learning rate of $1e-5$, a weight decay of $0.001$, and a batch size of 64 for optimization. Each batch is composed of $32$ randomly selected normal and abnormal videos. All experiments are carried out on a single 3090 GPU.

\subsection{Comparison with State-of-the-art Methods}

\begin{table}[t!]
	\centering
	\footnotesize
	\renewcommand\arraystretch{1.12} 
	\setlength{\tabcolsep}{2.3mm}{
		\begin{tabular}{c | c |  c  | c}
			\toprule
			Supervision   &    Methods &   Feature & AP(\%) \\
			\midrule 
			\multirow{2}* {Semi} &OCSVM~\cite{scholkopf1999support} & I3D+VGGish & $27.25$ \\
			~ & Conv-AE ~\cite{hasan2016learning} & I3D+VGGish & $30.77$ \\
			\midrule
			\multirow{9}* {Weak} & Wu \emph{et al.}~\cite{wu2020not}& I3D & $75.41$ \\
			~ & CA-VAD~\cite{chang2021contrastive}&  I3D & $76.90$ \\
			~ & RTFM~\cite{tian2021weakly}&  I3D  & $77.81$ \\
			~ & WSTD~\cite{huang2022weakly}&  I3D & $74.60$ \\
			~ & MSL~\cite{li2022self}&  I3D  & $78.28$ \\
			~ & DDL~\cite{pu2022locality}& I3D & $80.72$ \\
			~ & MGFN~\cite{chen2023mgfn} & I3D & $79.19$ \\
			~ & CU-Net~\cite{zhang2023exploiting}& I3D & $78.74$ \\
			\rowcolor{gray!10}~ & \textbf{Ours} &  I3D  & \textbf{81.66} \\
			\cmidrule(r){1-4}
			\multirow{6}*{Weak} & SVM baseline & I3D+VGGish & $50.78$ \\
			~& MIL-Rank~\cite{sultani2018real} &  I3D+VGGish & $73.20$ \\
			~ & Wu \emph{et al.}~\cite{wu2020not}& I3D+VGGish & $78.64$ \\
			~ & Wu \emph{et al.}~\cite{wu2021learning}& I3D+VGGish & $75.90$ \\
			~ & CU-Net~\cite{zhang2023exploiting}& I3D+VGGish & $81.43$ \\
			\rowcolor{gray!10}~ & \textbf{Ours} &  I3D+VGGish  & \textbf{83.13} \\
			\bottomrule
	\end{tabular}}
	\caption{Results on the XD-Violence dataset.}
	\label{tab:Violence_Res}
\end{table}

\noindent \textbf{Evaluation on XD-Violence.} As in Tab.~\ref{tab:Violence_Res}, our method achieves favorable performance on the XD-Violence dataset. Specifically, our method with only I3D features sets a new state-of-the-art performance of $81.66$\% AP. With the fused features of I3D  and VGGish, our method surpasses the state-of-the-art model~\cite{zhang2023exploiting} with an absolute performance gain of $1.7$\% AP on the test set.
In addition, the comparison of results obtained by these two types of features shows that audio features (VGGish) can largely improve anomaly detection performance.

\begin{table}[t!]
	\centering
	\footnotesize
	\renewcommand\arraystretch{1.11} 
	\setlength{\tabcolsep}{2.3mm}{
		\begin{tabular}{c | c | c | c}
			\toprule
			Supervision   &    Methods &   Feature & AUC(\%) \\
			\midrule 
			\multirow{2}* {Semi} & Luo \emph{et al.}~\cite{luo2017revisit} &   TSN & $57.89$ \\
			~ & Liu \emph{et al.}~\cite{liu2018future} &   - & $69.13$ \\
			\midrule
			\multirow{9}* {Weak} & MIL-Rank~\cite{sultani2018real} &  C3D & $83.27$ \\
			~ & MIL-Rank~\cite{sultani2018real} &  TSN & $85.95$ \\
			~ & MIL-Rank~\cite{sultani2018real} &  I3D & $88.34$ \\
			~ & Motion-Aware~\cite{zhu2019motion}&   TSN & $83.08$ \\
			~ & MIST~\cite{feng2021mist}&  I3D & $89.26$ \\
			~ & WSAL~\cite{lv2021localizing}&   TSN & $89.64$ \\
			~ & RTFM~\cite{tian2021weakly}&  I3D & $89.64$ \\
			~ & CU-Net~\cite{zhang2023exploiting}& I3D & $91.66$ \\
			\rowcolor{gray!10}~ & \textbf{Ours} &  I3D & \textbf{93.10} \\
			\bottomrule
	\end{tabular}}
	\caption{Results on the TAD dataset.}
	\label{tab:TAD_Res}
\end{table}

\begin{table}[t!]
	\footnotesize
	\centering
	\renewcommand\arraystretch{1.1} 
	\setlength{\tabcolsep}{2.1mm}{\begin{tabular}{c | c | c | c}
			\toprule
			Supervision   &    Methods &   Feature & AUC(\%) \\
			\midrule
			\multirow{4}* {Semi} & Lu \emph{et al.}~\cite{lu2013abnormal} & Dictionary & $65.51$ \\
			~ & Conv-AE~\cite{hasan2016learning} &  AE & $50.60$ \\
			~ & Object-Centric~\cite{ionescu2019object} &   - & $61.60$ \\
			~ & Sun \emph{et al.}~\cite{sun2020scene} &    - & $72.70$ \\
			\midrule
			\multirow{19}* {Weak} & Binary classiﬁer  &  C3D & $50.00$ \\
			~ & MIL-Rank~\cite{sultani2018real} &  C3D & $75.41$ \\
			~ & IBL~\cite{zhang2019temporal}&  C3D &  $78.66$ \\
			~ & Motion-Aware~\cite{zhu2019motion}&   AE  & $79.00$ \\
			~ & GCN~\cite{zhong2019graph}&   TSN  & $82.12$ \\
			~ & Wu \emph{et al.}~\cite{wu2020not}& I3D & $82.44$ \\
			~ & CLAWS~\cite{zaheer2020claws}&   C3D & $83.03$ \\
			~ & MIST~\cite{feng2021mist}&  I3D & $82.30$ \\
			~ & RTFM~\cite{tian2021weakly}&  I3D & $84.30$ \\
			~ & CA-VAD~\cite{chang2021contrastive}&  I3D & $84.62$ \\
			~ & WSAL~\cite{lv2021localizing}&  TSN & $85.38$ \\
			~ & Wu \emph{et al.}~\cite{wu2021learning}& I3D & $84.89$ \\
			~ & BN-SVP~\cite{sapkota2022bayesian}&  I3D  & $83.39$ \\
			~ & WSTD~\cite{huang2022weakly}&  I3D & $84.04$ \\
			~ & DDL~\cite{pu2022locality}& I3D & $85.12$ \\
			~ & MSL~\cite{li2022self}&  I3D  & $85.30$ \\
			~ & MGFN*~\cite{chen2023mgfn} & I3D & $83.45$ \\
			~ & CU-Net~\cite{zhang2023exploiting}&  I3D & $86.22$ \\
			\rowcolor{gray!10}~ & \textbf{Ours} &  I3D   & \textbf{86.33} \\
			\bottomrule
	\end{tabular}}
	\caption{Results on the UCF-Crime dataset. * denotes reproduced results using the official code.} 
	\label{tab:UCF_Res}
\end{table}

\vspace{1mm}
\noindent \textbf{Evaluation on TAD.} Tab.~\ref{tab:TAD_Res} compares the results of our methods with existing semi-supervised and weakly supervised approaches on the TAD dataset. 
Our method achieves significant improvement.
In particular,
our method achieves clear improvements against the state-of-the-art model~\cite{zhang2023exploiting} by 1.44\% in terms of AUC.

\vspace{1mm}
\noindent \textbf{Evaluation on UCF-Crime.} The performance comparisons with state-of-the-art weakly supervised methods and several semi-supervised methods on the UCF-Crime dataset are shown in Tab.~\ref{tab:UCF_Res}. 
It shows that our method achieves the highest AUC.
We note that the performance improvement on this dataset is not as significant as that on the XD-Violence and TAD datasets. 
The reason could be that most of the anomalies in this dataset are continuous video segments. Since the videos in UCF-Crime are shot by a fixed camera, the difference between these continuous abnormal segments in the same video is minimal. 
As a result, existing methods tend to generate similar anomaly scores on the most anomalous segment and the sub-anomalous segments, that is, the predicted anomaly scores on most of the abnormal segments are very high.
Therefore, the performance gain of the proposed method is not as apparent as that on the other two datasets.

\subsection{Ablation Study}

In this subsection,
we further validate the effectiveness of each design. Experimental results are obtained on the XD-Violence dataset, which is the largest and the most challenging benchmark.

\begin{table}[t]
	\centering
	\footnotesize
	\renewcommand\arraystretch{1.11} 
	\setlength{\tabcolsep}{3.2mm}{
		\begin{tabular}{c|c|c|c|c|c}
			\toprule
			\#&Baseline &MSTM& DA& EO & AP(\%) \\
			\midrule
			\footnotesize{\textcircled{\scriptsize{1}}} &\checkmark  &  ~ & ~ & ~ & $79.49$ \\
			\footnotesize{\textcircled{\scriptsize{2}}} &\checkmark  & \checkmark & ~ & ~ & $80.93$ \\
			\footnotesize{\textcircled{\scriptsize{3}}} &\checkmark  & \checkmark  & ~  &\checkmark & $82.55$ \\
			\rowcolor{gray!10}\footnotesize{\textcircled{\scriptsize{4}}} &\checkmark  & \checkmark  & \checkmark  & \checkmark  &  \textbf{83.13} \\
			\bottomrule
	\end{tabular}}
	\caption{Ablation study on the XD-Violence dataset. MSTM denotes the Multi-Scale Temporal Modeling, DA represents the Dynamic Assessment, and EO stands for the Erasure Operation. }
	\label{tab:Ablation_Res}
\end{table}

\begin{table}[t]
	\centering
	\footnotesize
	\renewcommand\arraystretch{1.25} 
	\setlength{\tabcolsep}{3.6mm}{
		\begin{tabular}{c | c | c | c }
			\toprule
			Methods   &  RTFM~\cite{tian2021weakly}  &   MS-TCT~\cite{dai2022ms} &  \cellcolor{gray!10}Ours  \\
			\midrule
			AP(\%) & $80.91$ & $81.66$ & \cellcolor{gray!10}\textbf{83.13} \\
			\bottomrule
	\end{tabular}}
	\caption{Comparison of different multi-scale temporal modeling methods on the XD-Violence dataset.}
	\label{tab:MSTM}
\end{table} 

\vspace{1mm}
\noindent \textbf{Effect of Multi-Scale Temporal Modeling.}  
The results are present in Tab.~\ref{tab:Ablation_Res}.
Comparing using our proposed MSTM (\textcircled{\footnotesize{2}}) with the baseline model (\textcircled{\footnotesize{1}}), it shows that the MSTM design brings $1.44$\% improvement in terms of AP. 
We also compare our MSTM module with the multi-scale network in RTFM~\cite{tian2021weakly} and MS-TCT~\cite{dai2022ms} respectively and report the results in Tab.~\ref{tab:MSTM}. 
Compared with RTFM~\cite{tian2021weakly}, MSTM introduces multi-scale global modeling and achieves an absolute gain of $2.22$\% in terms of AP. 
For MS-TCT~\cite{dai2022ms}, which is tailored to densely-labeled action detection, it exploits contextual relations at different scales to model the regularity of action patterns. 
However, anomalies are often irregular events and lack predictable patterns. Therefore, the performance of the multi-scale network used in MS-TCT~\cite{dai2022ms} is not as good as ours.

\vspace{1mm}
\noindent \textbf{Effect of Dynamic Erasing.} We also evaluate the effect of the dynamic erasing strategy consisting of DA and EO. With the dynamic erasing strategy (DE), the network can continuously mine different abnormal segments to achieve complete anomaly detection. 
As shown in Tab.~\ref{tab:Ablation_Res}, when we incorporate the dynamic erasing strategy (\textcircled{\footnotesize{4}}), the performance increases from $80.93$\% to $83.13$\%, achieving an absolute gain of $2.2$\% compared to without using DE (\textcircled{\footnotesize{2}}). 
It demonstrates that mining complete abnormal segments significantly improves the performance.
Besides, we replace the proposed dynamic erasing strategy with a static erasing strategy, which performs the erasure operation for all videos (\emph{i.e.}, without DA). It shows that without DA, the AP drops from $83.13$\% to $82.55$\% (\textcircled{\footnotesize{4}} vs \textcircled{\footnotesize{3}}), demonstrating the effectiveness of dynamically assessing whether a video needs to erase the prominent anomalous segments.

\subsection{Evaluation on Hyper-parameters}

\begin{figure}[!t]
	\centering
	\includegraphics[width=\linewidth,trim=5 14 0 0,clip]{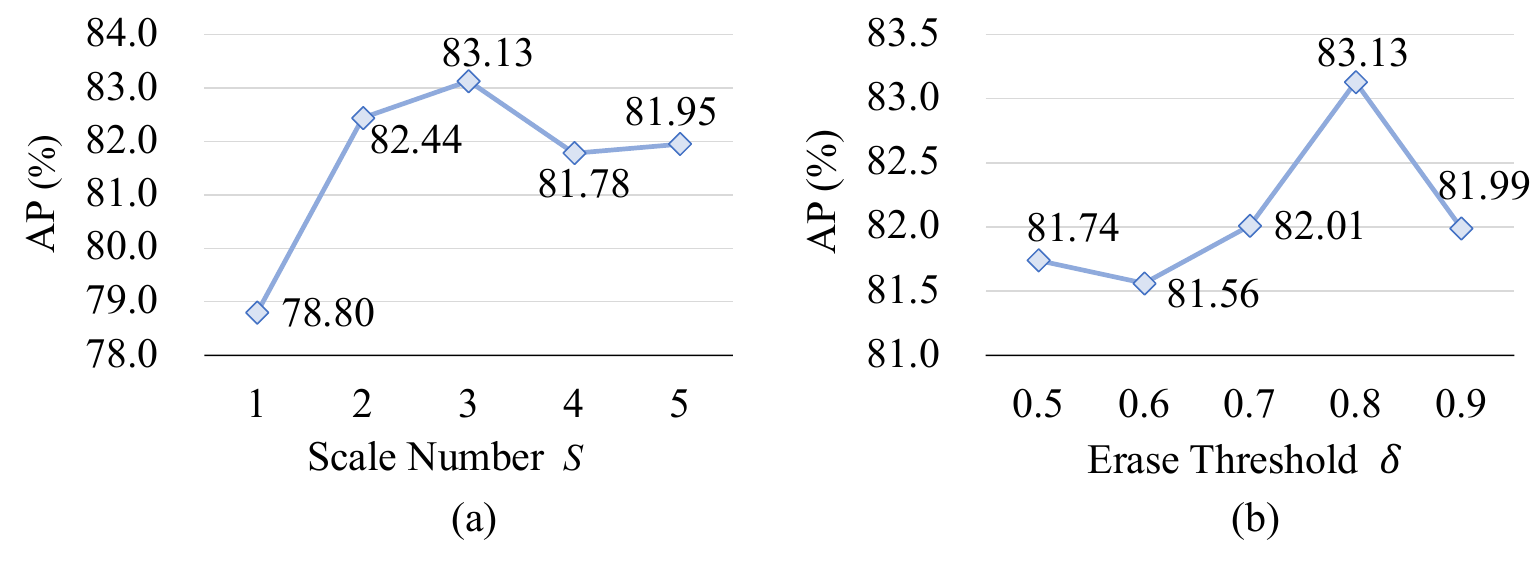}
	\vspace{-5mm}
	\caption{
		Comparison of different hyper-parameter settings.
	}
	\label{fig:parameter}
\end{figure}

In this subsection, we evaluate the effect of two important hyper-parameters using the XD-Violence dataset, namely the number of scales $S$ and the erase threshold $\delta$.

\vspace{1mm}
\noindent \textbf{Scale Number $S$.} We first  explore the influence of the scale number $S$, as shown in Fig.~\ref{fig:parameter}(a). The performance consistently increases as $S$ increases from 1 to 3. But when $S>3$, it does not yield extra performance gains. The reason could be that an excessive number of scales bring redundant information and slightly impair the performance.

\vspace{1mm}
\noindent \textbf{Erase Threshold $\delta$.} In Fig.~\ref{fig:parameter}(b), we delve deeper into the erase threshold $\delta$. Since the dividing threshold score between abnormal and normal is usually regarded as 0.5, we select scores between 0.5 and 0.9 as different thresholds. 
It shows that over-high or over-low $\delta$ cannot lead to accurate anomaly detection results. When $\delta$ is set too low, it may erase too many abnormal segments, causing the network to focus on normal segments instead. Conversely, if $\delta$ is set too high, it may fail to detect substantial abnormal segments completely.

\subsection{Behavior of Erasing Strategy}

\begin{figure}[!t]
	\centering
	\includegraphics[width=\linewidth,trim=5 3 0 0,clip]{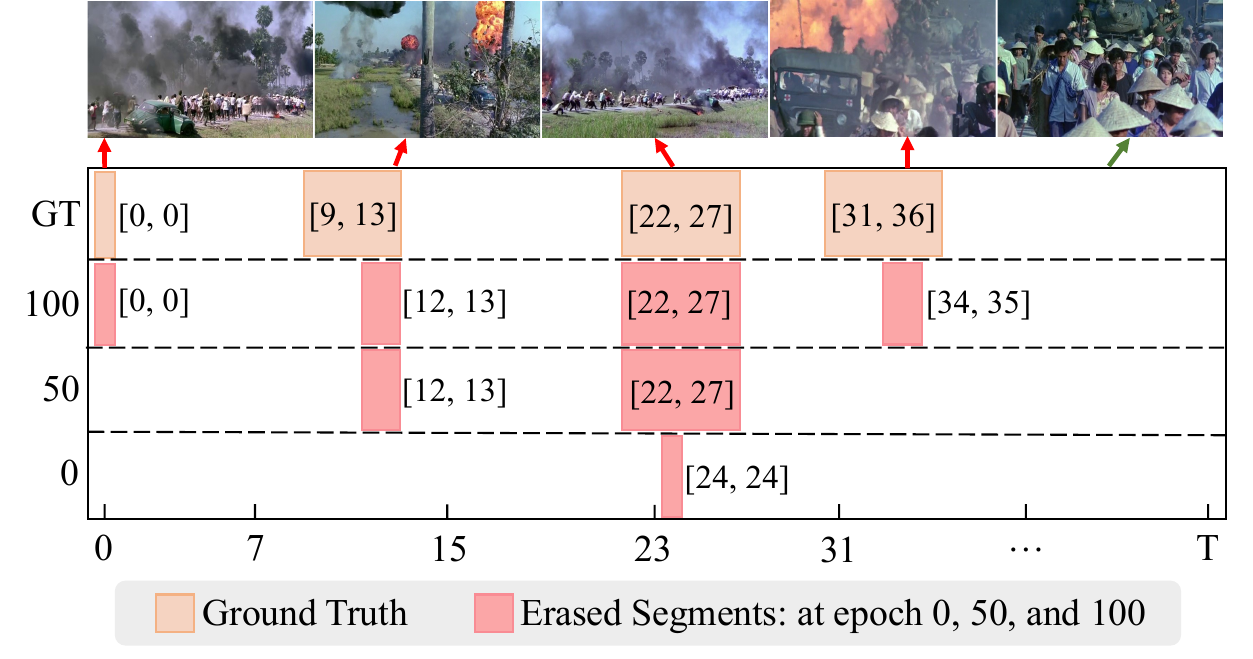}
	\vspace{-5mm}
	\caption{
		Visualization of the erasing process.
	}
	\label{fig:erase_process}
\end{figure}
Fig.~\ref{fig:erase_process} shows the erasure results on one video from the XD-Violence training split. 
Because the frame-level anomaly annotation is not provided, we manually annotate the ground truth of this video to facilitate the visualization of the erasing process.
As training epochs increase, our method gradually erases more abnormal segments. This proves that our erasing strategy enables the model to progressively learn to detect more abnormal segments, effectively addressing the completeness issue of anomaly detection.

 \begin{figure}[!t]
	\centering
	
	\subcaptionbox{Shooting\label{fig:Shooting}}{\includegraphics[width=\mysize]{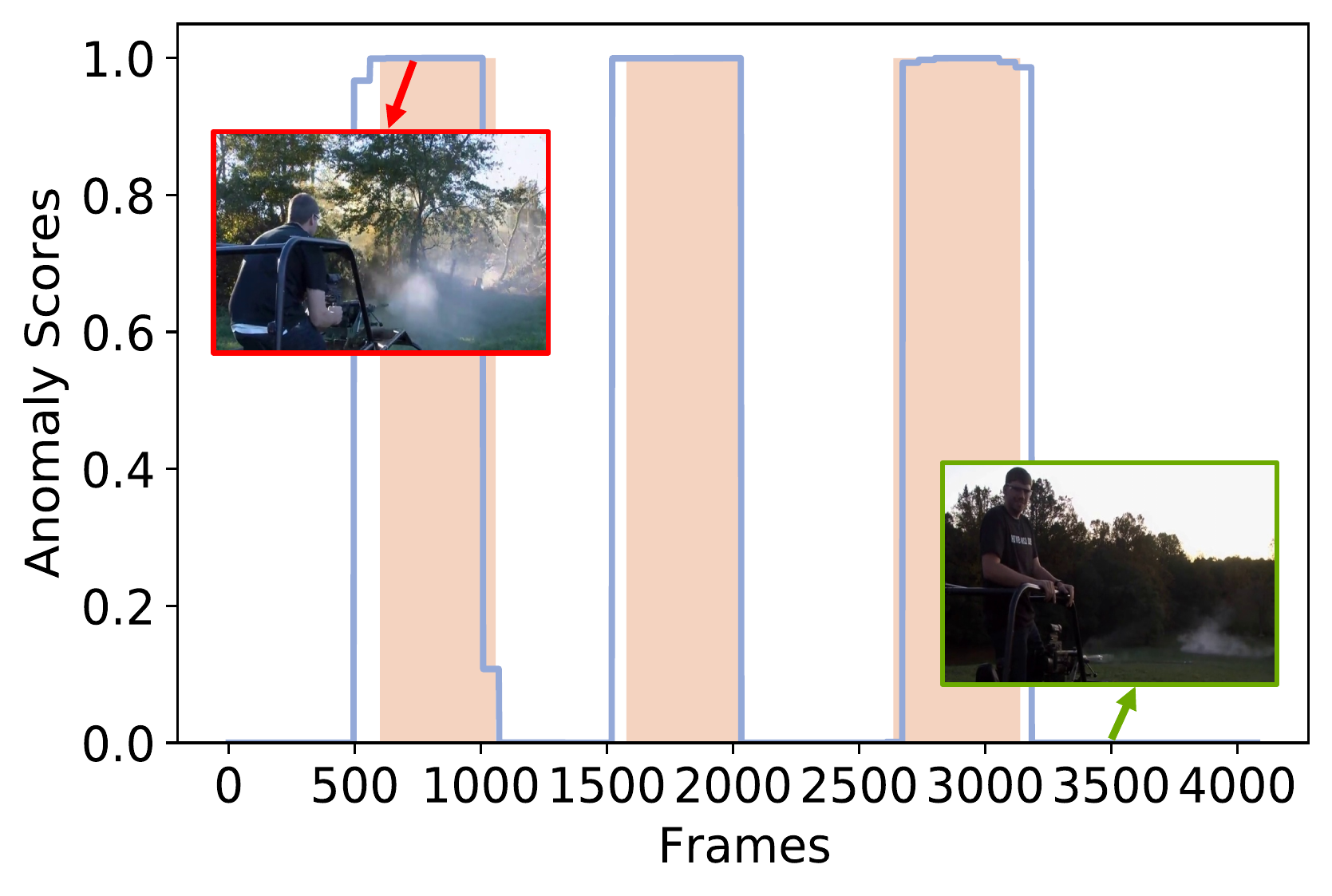}}
	\subcaptionbox{Fighting\label{fig:Fighting}}{\includegraphics[width=\mysize]{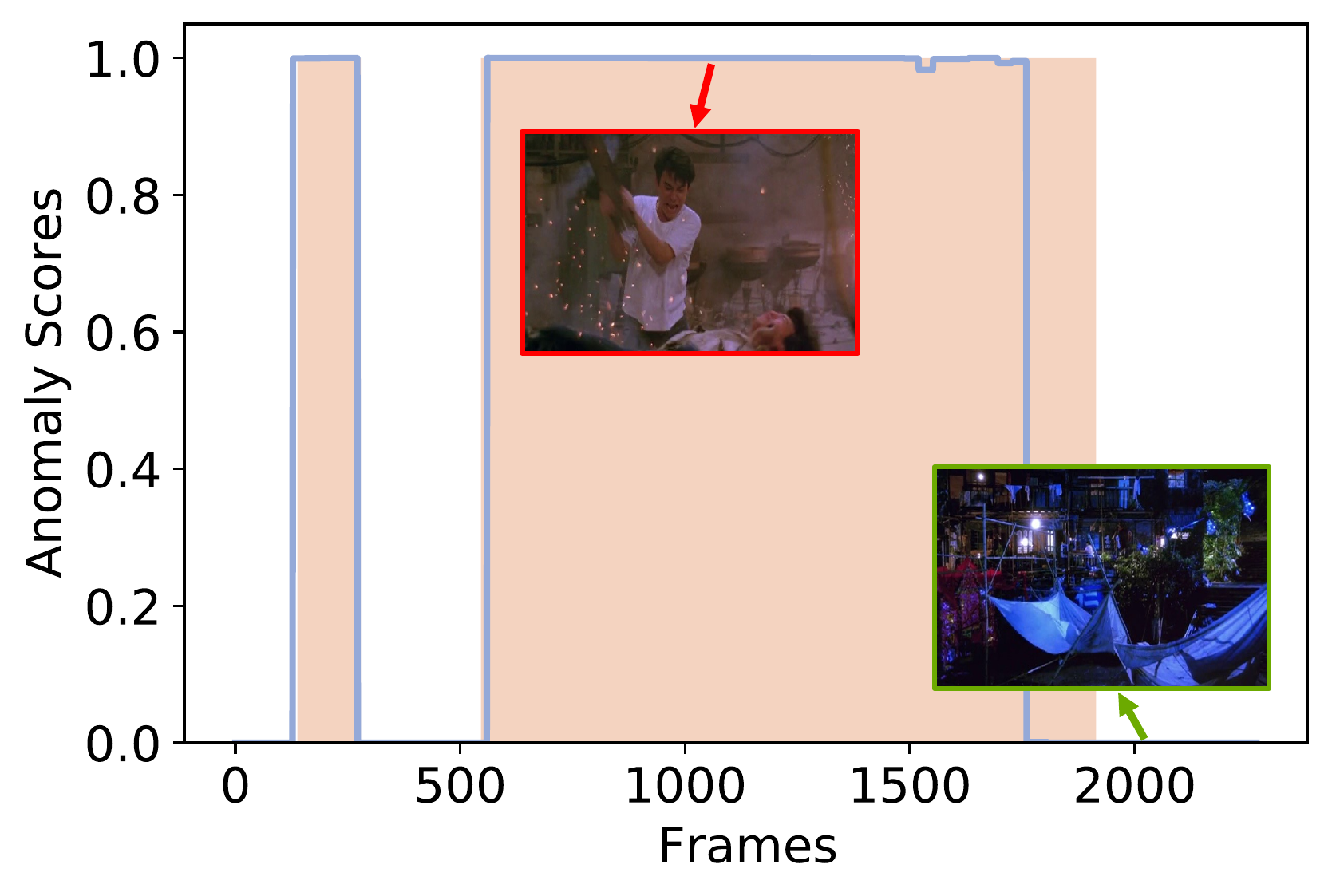}}
	\subcaptionbox{Riot\label{fig:Riot}}{\includegraphics[width=\mysize]{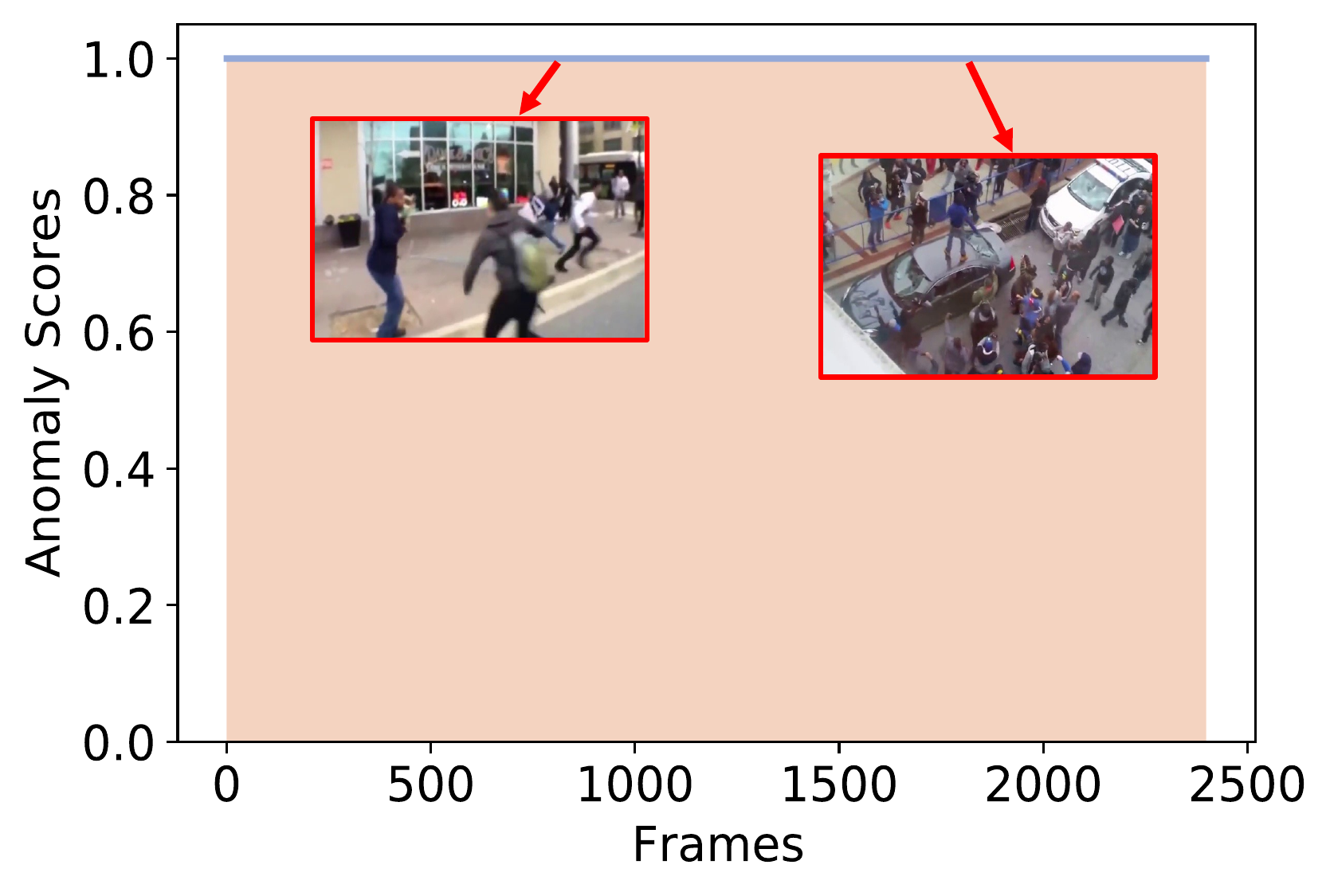}}
	\subcaptionbox{Normal\label{fig:violence_nor}}{\includegraphics[width=\mysize]{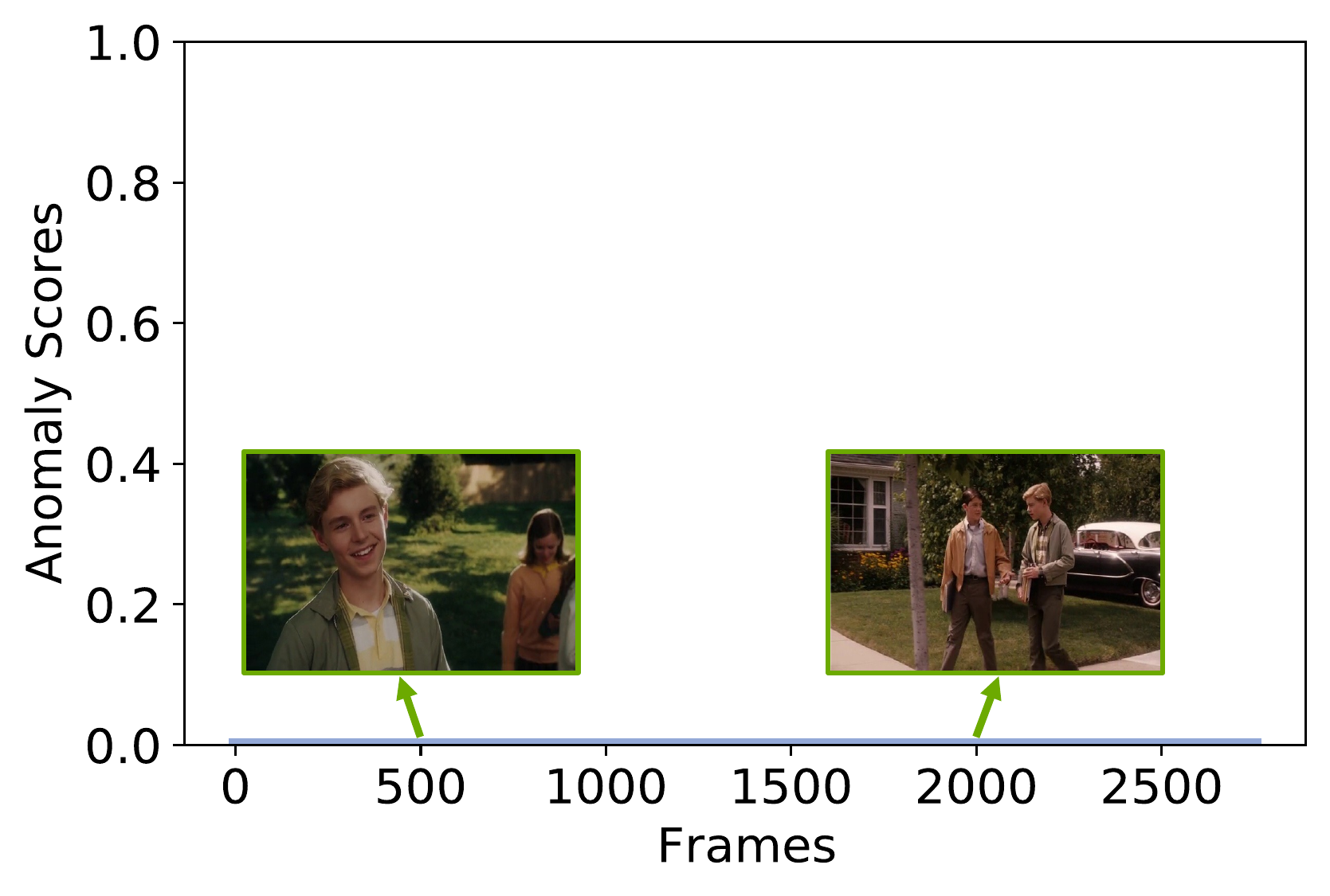}}
	\vspace{-1mm}
	\caption{Qualitative results on XD-Violence test videos. The orange square area represents the ground truth of anomalies, and the blue curve depicts the predictions for each video frame.}
	\label{fig:anomaly_score}
\end{figure}

\subsection{Qualitative Results}

We present several qualitative results of our approach on the XD-Violence test set in Fig.~\ref{fig:anomaly_score}. 
In Fig.~\ref{fig:anomaly_score}(a), although there are three discontinuous short-term anomalies (Shooting), our method accurately detects all the anomalies. Similarly, as demonstrated in Fig.~\ref{fig:anomaly_score}(b), our method predicts higher anomaly scores for non-continuous long and short anomalies (Fighting) in a video. Fig.~\ref{fig:anomaly_score}(c) depicts a video full of anomalies (Riot), and our method can detect all anomalies. 
For normal videos, as depicted in Fig.~\ref{fig:anomaly_score}(d), the anomaly scores predicted by our method are consistently close to 0. For anomalies of different types and durations, our method consistently provides accurate anomaly scores, demonstrating its effectiveness.

\section{Conclusion}

In this work, we propose a dynamic erasing network based on multi-scale temporal features for weakly supervised video anomaly detection. Our multi-scale temporal modeling module is designed to capture anomalies of different durations, which can enhance event representation.
Moreover, we design a dynamic erasing strategy that exploits the similarity between segments to dynamically assess the completeness of detected anomalies and selectively erase the prominent segments, forcing the model to detect gentle abnormal segments.
Extensive experimental results on three anomaly detection datasets demonstrate the effectiveness of our approach.
{
    \small
    \bibliographystyle{ieeenat_fullname}
    \bibliography{main}
}

\end{document}